\documentclass[11pt,a4paper]{article}
\usepackage[hyperref]{acl2019}
\usepackage{times}
\usepackage{latexsym}
\usepackage{graphicx}
\usepackage{multirow}
\usepackage{xcolor,colortbl}
\usepackage{easyReview}
\usepackage{booktabs}
\usepackage{amssymb}
\newcommand{\lgray}{\cellcolor{lightgray!25}}
\newcommand{\rood}[1]{} 

\usepackage{url}

\aclfinalcopy 

\definecolor{jade}{rgb}{0.0, 0.66, 0.42}
\newcommand{\ipr}[2]{\underline{#1}\textsubscript{\tiny{#2}}}
\newcommand{\ripr}[2]{\textcolor{red}{\underline{#1}\textsubscript{\tiny{#2}}}}
\newcommand{\gipr}[2]{\textcolor{jade}{\underline{#1}\textsubscript{\tiny{#2}}}}
\newcommand{\gripr}[2]{\textcolor{gray}{\underline{#1}\textsubscript{\tiny{#2}}}}
\newcommand{\rc}[1]{\textcolor{red}{#1}}
\newcommand{\gc}[1]{\textcolor{jade}{#1}}

\title{Adapt or Get Left Behind:\\
Domain Adaptation through BERT Language Model Finetuning for Aspect-Target Sentiment Classification}

\author{Alexander Rietzler, Sebastian Stabinger, Paul Opitz, Stefan Engl \\
  DeepOpinion.ai at Innsbruck, Austria \\
  \texttt{ \{firstname.lastname\}@deepopinion.ai}
}
\date{}

\begin{document}
\maketitle
\begin{abstract}  
  \rood{Motivation where in Real Life your contribution is relevant. Inspiration on method. 2 Sentences}
  Aspect-Target Sentiment Classification (ATSC) is a subtask of Aspect-Based Sentiment Analysis (ABSA), which has many applications e.g. in e-commerce, where data and insights from reviews can be leveraged to create value for businesses and customers.
  Recently, deep transfer-learning methods have been applied successfully to a myriad of Natural Language Processing (NLP) tasks, including ATSC.
  Building on top of the prominent BERT language model, we approach ATSC using a two-step procedure: self-supervised domain-specific BERT language model finetuning, followed by supervised task-specific finetuning.
  Our findings on how to best exploit domain-specific language model finetuning enable us to produce new state-of-the-art performance on the SemEval 2014 Task 4 restaurants dataset.
  In addition, to explore the real-world robustness of our models, we perform cross-domain evaluation.
  We show that a cross-domain adapted BERT language model performs significantly better than strong baseline models like vanilla BERT-base and XLNet-base. Finally, we conduct a case study to interpret model prediction errors.
\end{abstract}

\section{Introduction}

\rood{
  Outcome: The Reader should have.... \\ \\
  The introduction can be framed 
  as such:
  the main contribution is 
  to find out the influences of cross domain finetuning, pre-training
  on both end-task performance and qualitativley on error analysis.
  And also how exactly the number of examples in lm-finetuning seen influence the end-task performance.
  \\
  Structure:
  1. Motivation: The absa task is a important task for application x/y \\
  2. Example:  \\
}

Sentiment Analysis (SA) is an active field of research in Natural Language Processing and deals with opinions in text. A typical application of classical SA in an industrial setting would be to classify a document like a product review into \textit{positive, negative} or \textit{neutral} sentiment polarity.

In constrast to SA, the more fine-grained task of Aspect Based Sentiment Analysis (ABSA)~\cite{Hu2004, Pontiki2014} aims to find both the aspect of an entity like a restaurant, and the sentiment associated with this aspect.

It is important to note that ABSA comes in two variants. We will use the sentence \textit{``I love their \underline{dumplings}''} to explain these variants in detail.

Both variants are implemented as a two-step procedure. The first variant is comprised of Aspect-Category Detection (ACD) followed by Aspect-Category Sentiment Classification (ACSC).
ACD is a multilabel classification task, where a sentence can be associated with a set of predefined aspect categories like \textit{``food''} and \textit{``service''} in the restaurants domain.
In the second step, ACSC, the sentiment polarity associated to the aspect-category is classified. For our example-sentence the correct result is the tuple \textit{(``food'', ``positive'')}.

The second variant consists of Aspect-Target Extraction (ATE) followed by Aspect-Target Sentiment Classification (ATSC).
ATE is a sequence labeling task, where terms like \textit{``dumplings''} are detected. In the second step, ATSC, the sentiment polarity associated to the aspect-target is determined. In our example the correct result is \textit{("dumplings", "positive")}.

In this paper, we focus on ATSC. 
In recent years, specialized neural architectures~\cite{Tang2015, Tang2016} have been developed that substantially improved modeling of this target-context relationship.
More recently, the Natural Language Processing community experienced a substantial shift towards using pre-trained language models~\cite{Peters2018, Radford2018, Howard, Devlin2019} as a base for many down-stream tasks, including ABSA~\cite{ArxSong, Xu2019, Sun2019}.
We still see huge potential that comes with this trend, which is why we approach the ATSC task using the BERT architecture.

As shown by \citet{Xu2019}, for the ATSC task the performance of models that were pre-trained on general text corpora is improved substantially by finetuning the language model on domain-specific corpora --- in their case review corpora --- that have not been used for pre-training BERT, or other language models.

We extend the work by Xu et al. by further investigating the behavior of finetuning the BERT language model in relation to ATSC performance.
In particular, our contributions are:

\begin{enumerate}
\item Analysis of the influence of the amount of training-steps used for BERT language model finetuning on the Aspect-Target Sentiment Classification performance.
\item Findings on how exploiting BERT language model finetuning enables us to achieve new state-of-the-art performance on the SemEval 2014 restaurants dataset.
\item Analysis of cross-domain adaptation between the laptops and restaurants domains. Adaptation is tested by self-supervised finetuning of the BERT language model on the target-domain and then supervised training on the ATSC task in the source-domain.
  In addition, the performance of training on the combination of both datasets is measured. 
\end{enumerate}

\section{Related Works}

We separate our discussion of related work into two areas: first, neural methods applied to ATSC that have improved performance solely by model architecture improvements. Secondly, methods that additionally aim to transfer knowledge from semantically related tasks or domains.

\subsection*{Architecture Improvements for Aspect-Target Sentiment Classification}

The datasets typically used for Aspect-Target Sentiment Classification are the SemEval 2014 Task 4 datasets~\cite{Pontiki2014} for the restaurants and laptops domain. Both datasets have only a small number of training examples.
One common approach to compensate for insufficient training examples is to invent neural architectures that better model ATSC. 
For example, in the past a big leap in classification performance was achieved with the use of the Memory Network architecture~\cite{Tang2016}, which uses memory to remember context words and explicitly models attention over both the target word and context.
It was found that making full use of context words improves their model compared to previous models~\cite{Tang2015} that make use of left- and right-sided context independently.

\citet{ArxSong} proposed Attention Encoder Networks (AEN)\rood{which they test both with GloVe wordvectors as input embeddings as well as BERT embeddings (AEN-BERT).}, a modification to the 
transformer architecture. The authors split the Multi-Head Attention (MHA) layers into Intra-MHA and Inter-MHA layers in order to model target words and context differently, which results in a more lightweight model compared to the transformer architecture.

Another recent performance leap was achieved by \citet{ArxZhaoa2019}, who model dependencies between sentiment words explicitly in sentences with more than one aspect-target by using a graph convolutional neural network. They show that their architecture performs particularly well if multiple aspects are present in a sentence.

\subsection*{Knowledge Transfer for Aspect-Target Sentiment Classification Analysis}

One approach to compensate for insufficient training examples is to transfer
knowledge across domains or across similar tasks.
\rood{* General Text 
  * General Review Domain
  * General Restaurants Review}

\citet{Li} proposed Multi-Granularity Alignment Networks (MGAN). They use this architecture to transfer knowledge from both an aspect-category classification task and also across different domains. They built a large scale aspect-category dataset specifically for this. 

\citet{He} transfer knowledge from a document-level sentiment classification task trained on the Amazon review dataset~\cite{He2016}.
They successfully apply pre-training by reusing the weights of a Long Short Term Memory (LSTM) network~\cite{Hochreiter1997} that has been trained on the document-level sentiment task. In addition, they apply multi-task learning where aspect and document-level tasks are learned simultaneously by minimizing a joint loss function.

Similarly, \citet{Xu2019} introduce a multi-task loss function to simultaneously optimize on BERT model's~\cite{Devlin2019} pre-training objectives as well as a question answering task.

In contrast to the methods described above that aim to transfer knowledge from 
a different source task - like question answering or document-level sentiment classification - this paper aims at transferring knowledge across different domains by self-supervised finetuning of the BERT language model.

\section{Methodology}
\rood{Outcome: Overview of the Subsections. Reader Should understand
  that in order to train our task we use the pipeline of using pre-trained
  bert, than language model finetuning and training in the mode for aspect based sentiment analysis. Tell that this is not new and has been 
  used in posttraining paper and also in BERT-pair paper.
  Tell the novel thing is that we research the influence of number of iterations,
  how out of domain generalization error is influenced by the LM-domain.
  Keep the first three subsections short, because its not our contribution.
}
We approach the Aspect-Target Sentiment Classification task using 
a two-step procedure. We use the pre-trained BERT architecture as a basis. In the first step we finetune the pre-trained weights of the language model further in a self-supervised way on a domain-specific corpus. In the second step we train the finetuned language model in a supervised way on the ATSC end-task. 

In the following subsections, we discuss the BERT architecture, how we finetune the language model, and how we transform the ATSC task into a BERT sequence-pair classification task~\cite{Sun2019}.
Subsequently, we discuss the different end-task training and domain-specific finetuning combinations
we employ to evaluate our model's generalization performance not only in-domain but also cross-domain.

Finally, we describe how we apply input reduction, an interpretation method for neural NLP models, to the ATSC task.
\subsection{BERT}
\rood{Outcome: Readers should understand that we use bert-base and 
  use the pre-trained weights and its architecture}

The BERT model builds on many previous innovations: contextualized word representations~\cite{Peters2018}, the transformer architecture~\cite{Vaswani2017}, and pre-training on a language modeling task with subsequent end-to-end finetuning on a downstream task~\cite{Radford2018, Howard}.
Due to being deeply bidirectional, the BERT architecture creates powerful sequence representations that perform extremely well on many downstream tasks~\cite{Devlin2019}.

The main innovation of BERT is that instead of using the objective of next-word prediction, a different objective is used to train the language model.
This objective consists of two parts.

The first part is the masked language model objective, where the model learns to predict randomly masked tokens from their context.

The second part is the next-sequence prediction objective, where the model needs to predict if a sequence $B$ would naturally follow the previous sequence $A$.
This objective enables the model to capture long-term dependencies better.
Both objectives are discussed in more detail in the next section.

As a base for our experiments we use the BERT\textsubscript{BASE} model, which has been pre-trained by the Google research team. It has the following parameters:
12 layers, 768 hidden dimensions per token and 12 attention heads. It has 110 million parameters in total.

For finetuning the BERT language model on a specific domain we use the weights of BERT\textsubscript{BASE} as a starting point.
\rood{For ATSC, an additional fully-connected layer on top of BERT with a softmax activation function is added.}

\subsection{BERT Language Model Finetuning}
\rood{Outcome: Readers should understand that we use language model (LM) finetuning
  (forget the 2 methods?!) as an intermediate step before classification
  Also they should understand that we try to optimize the number of iteration compared to previous work we would know the limit of significance of improvement.
}
As the first step of our procedure we perform language model finetuning of the BERT
model using domain-specific corpora. 
Algorithmically, this is equivalent to pre-training.
The domain-specific language model finetuning as an intermediate step to ATSC has been described by \citet{Xu2019}. As an extension to their paper we investigate the limits of language model finetuning in terms of how end-task performance is dependent on the amount of training steps. 

The training input representation for language model finetuning consists of two sequences $s_A$ and $s_B$ in the format of $``\textrm{[CLS]} \ s_{A} \ \textrm{[SEP]} \ s_{B} \ \textrm{[SEP]}"$,
where 
\textrm{[CLS]} is a dummy token used for downstream classification and \textrm{[SEP]} are separator tokens.
\subsubsection*{Masked Language Model Objective}
The sequences $A$ and $B$ have tokens randomly masked out in order for the model to learn to predict them. 
The following example shows how domain-specific finetuning could alleviate the 
bias from pre-training on a Wikipedia corpus:
\textit{``The touchscreen is an} [MASK] \textit{device''}. In the fact-based context of Wikipedia the [MASK] could be \textit{``input''} and in the review domain a typical guess could be the general opinion word \textit{``amazing''}.
\subsubsection*{Next-Sentence Prediction}
In order to train BERT to capture long-term dependencies better, the
model is trained to predict whether sequence $B$ follows sequence $A$. If
this is the case, sequence A and sequence B are jointly sampled from
the same document in the order they appear naturally. Otherwise
the sequences are sampled randomly from the training corpus.

\subsection{Aspect-Target Sentiment Classification}
\rood{Outcome: Readers should understand how we use a linear pooling layer on top of bert and that has been used by 2 other papers in a similar manner.
}

The ATSC task aims at classifying sentiment polarity into the three classes \textit{positive, negative, neutral} with respect to an aspect-target.
The input to the classifier is a tokenized sentence $s=s_{1:n}$ and a target $t=s_{j:j+m}$ contained in the sentence, where $j < j+m \leq n$.
Similar to previous work by \citet{Sun2019}, we transform the input into a format compatible with BERT sequence-pair classification tasks:
$``\textrm{[CLS]} \ s \ \textrm{[SEP]} \ t \ \textrm{[SEP]}"$.

In the BERT architecture the position of the token embeddings is structurally maintained after each Multi-Head Attention layer. Therefore, we refer to the last hidden representation of the [CLS] token as $h_{[CLS]} \in \mathbf{R}^{768 \times 1}$.
The number of sentiment polarity classes is three. A distribution $p \in [0,1]^3$ over these classes is predicted using a fully-connected layer with 3 output neurons on top of $h_{[CLS]}$, followed by a softmax activation function
\[ p = \textrm{softmax}(W \cdot h_{[CLS]} + b)\textrm{,} \]
where $b \in \mathbf{R}^3$ and $W \in \mathbf{R}^{3 \times 768}$.
Cross-entropy is used as the training loss. The way we use BERT for classifying the sentiment polaritites is equivalent to how BERT is used for sequence-pair classification tasks in the original paper~\cite{Devlin2019}.

\subsection{Domain Adaptation through Language Model Finetuning}
\rood{Outcome: Reader should understand in detail that we train on many variants
  in order to check performance}
In academia, it is common that the performance of a machine learning model is evaluated \textit{in-domain}. This means that the model is evaluated on a test set that comes from the same distribution as the training set. In real-world applications this setting is not always valid, as the trained model is used to predict previously unseen data.

In order to evaluate the performance of a machine learning model more robustly,
its generalization error can be evaluated across different domains, i.e. \textit{cross-domain}. To optimize cross-domain performance, the model itself can be adapted towards a target domain. This procedure is known as Domain Adaptation, which is a special case of Transductive Transfer Learning in the taxonomy of \citet{Ruder2019}. Here, it is typically assumed that supervised data for a specific task is only available for a \textit{source domain} $S$, whereas only unsupervised data is available in the \textit{target domain} $T$. The goal is to optimize performance of the task in the target domain while transferring task-specific knowledge from the source domain.

If we map this framework to our challenge, we define Aspect-Target Sentiment Classification as the transfer-task and BERT language model finetuning is used for domain adaptation. In terms of which domain is finetuned on, the full transfer-procedure can be expressed in the following way:

\[ D_{LM} \rightarrow D_{Train} \rightarrow D_{Test}. \]

Here, $D_{LM}$ stands for the domain on which the language model is finetuned and can take on the values of \textit{Restaurants}, \textit{Laptops} or \textit{(Restaurants $\cup$ Laptops)}.
The domain for training $D_{Train}$ can take on the same values; for the joint case the training datasets for laptops and restaurants are simply combined.
The domain for testing $D_{Test}$ can only take the value \textit{Restaurants} or \textit{Laptops}.
%

Combining finetuning and training steps gives us nine different evaluation scenarios, which we group into the following four categories: 

\subsection*{In-Domain Training} 
ATSC is trained on a domain-specific dataset and evaluated on the test set from the same domain. This can be expressed as \\$D_{LM} \rightarrow T \rightarrow T,$ where $T$ is our target domain and can be either \textit{Laptops} or \textit{Restaurants}. It is expected that the performance of the model is highest if $D_{LM} = T$.

\subsection*{Cross-Domain Training}
ATSC is trained on a domain-specific dataset and evaluated on the test set from the other domain. This can be expressed as \\$D_{LM} \rightarrow S \rightarrow T,$ where $S\neq T$ are source and target domain and can be either \textit{Laptops} or \textit{Restaurants}.

\subsection*{Cross-Domain Adaptation}
As a special case of cross-domain training we expect performance to be optimal if $D_{LM} = T$. This is the variant of \textit{Domain Adaptation} and is written as \\
  $T \rightarrow S \rightarrow T.$
  
\subsection*{Joint-Domain Training} 
ATSC is trained on both domain-specific datasets jointly and evaluated on both test sets independently. This can be expressed as \\$D_{LM} \rightarrow (S \cup T) \rightarrow T,$ where $S\neq T$ are source- and target domain and can either be \textit{Laptops} or \textit{Restaurants}.

\subsection{Input Reduction for Model Interpretation}
\label{sect:Input-reduction}
Input reduction is an interpretation method for neural models introduced by \citet{Feng2018}, which tries to find a subset of the most important words of a document that contribute most to a prediction.

We use this interpretation method to illustrate the predictions of our models on the test set in order to find causes for classification errors, and also to find qualitative differences between our models and baseline models.

The input reduction method resembles a process that iteratively removes unimportant words from the input while the model's  prediction is maintained. The idea is that the remaining set of words one iteration before the prediction flips are the most important ones. 
As pointed out by \citet{Feng2018} for this method to work, a machine learning model needs to compute meaningful confidence values for unseen input. 
For our task, we find empirically that the predicted probabilities computed for our test set examples work well enough as a confidence approximation, which means that most of the reduced input for the examples discussed in  \autoref{sect:Casestudy} allows for a meaningful interpretation.

Let $\textbf{x}=[x_1,x_2, \ldots x_n]$ be the input sentence represented as a list of tokens and $p(y|\textbf{x})$ the predicted probability of label $y$, and $y=\textrm{argmax}_{\hat{y}} \ p(\hat{y}|\textbf{x})$ the originally predicted label.
The importance of a word is defined as 

\[ g(x_i) = p(y|\textbf{x}) - p(y|\textbf{x}_{-i}). \]

Put differently, the importance of a word is the prediction probability towards the original label of a sentence containing the word minus the prediction probability of the sentence without the same word.

We apply this formula to iteratively remove the word with the lowest importance until the prediction changes to another label. Due to the nature of the ATSC task, we make an exception for words that are part of the aspect-target phrase, which we do not remove during an iteration. This allows us to maintain the context with respect to the aspect-target.

\section{Experiments}
In our experiments we aim to answer the following research questions (RQs):

RQ1: How does the number of training iterations in the BERT language model finetuning stage influence the ATSC end-task performance? At what point does performance start to improve, when does it converge?

RQ2: If trained in-domain, what ATSC end-task performance can be reached through fully exploited finetuning of the BERT language model?

RQ3: If trained cross-domain in the special case of domain adaptation, what ATSC end-task performance can be reached if BERT language model finetuning is fully exploited?

\subsection{Datasets for Classification and Language Model Finetuning}
\rood{Outcome: Readers understand how we make use/preprocess and so on of SemEval 2014 Datasets and the corpora for LM finetuning}

We conduct experiments using the two SemEval 2014 Task 4 Subtask 2
datasets\footnote{\url{http://alt.qcri.org/semeval2014/task4}}~\cite{Pontiki2014}
for the laptops and the restaurants domain. The two datasets contain
sentences with one or multiple marked aspect-targets that each have a 3-level sentiment polarity (\textit{positive, neutral
  or negative}) associated. In the original dataset the
\textit{conflict} class is also present. Here, the \textit{conflict} labels are dropped for reasons of comparability with \citet{Xu2019}. Detailed statistics for both datasets are shown in \autoref{tab:datasets}.

For BERT language model finetuning we prepare three corpora for the
two domains of laptops and restaurants. For the restaurants domain we
use Yelp Dataset Challenge
reviews\footnote{\url{https://www.yelp.com/dataset/challenge}} and
for the laptops domain we use Amazon Laptop reviews~\cite{He2016}. For
the laptop domain we filtered out reviews that appear in the
SemEval 2014 laptops dataset to avoid training bias for the test data.
To be compatible with the next-sentence prediction task used during
fine tuning, we removed reviews containing fewer than two sentences from the corpora.

For the laptop corpus, $1,007,209$ sentences are left after pre-processing.
For the restaurants domain, where more reviews are available, we sampled $10,000,000$
sentences to have a sufficient amount of data for fully exploited language model finetuning.
In order to compensate for the smaller amount of finetuning data in the laptops domain, we finetune for more epochs, 30 epochs in the case of the laptops domain compared to 3 epochs for the restaurants domain, so that the BERT model trains on about 30 million sentences in both cases. This means that a single sentence can appear multiple times with a different language model masking.

We also create a mixed corpus to jointly finetune on both domains. Here, we sample 1 million restaurant reviews and combine them with the laptop reviews. This results in about 2 million reviews that are finetuned for 15 epochs.
The exact statistics for the three finetuning corpora are shown in the top of \autoref{tab:datasets}.

We release code to reproduce generation of our finetuning corpora\footnote{\url{https://github.com/deepopinion/domain-adapted-atsc}}.

\begin{table}[h!]
  \begin{center}
    \scalebox{0.80}{
      \begin{tabular}{lllllll}
        \toprule
        \textbf{Corpus} & \multicolumn{3}{c}{\textbf{Sentences}} & \multicolumn{3}{c}{\textbf{Finetuning Epochs}} \\
        \midrule
        Laptops & \multicolumn{3}{c}{1,007,209} & \multicolumn{3}{c}{30} \\
        Restaurants & 
                      \multicolumn{3}{c}{10,000,000} &
                                                       \multicolumn{3}{c}{3}
        \\
        Lapt.+Rest. &
                      \multicolumn{3}{c}{2,007,213} &
                                                      \multicolumn{3}{c}{15} \\
        \midrule[.08em]
        \multirow{2}{*}{\textbf{Dataset}}
                        & \multicolumn{2}{c}{\textbf{Positive}} 
                                                                 & \multicolumn{2}{c}{\textbf{Negative}} 
                & \multicolumn{2}{c}{\textbf{Neutral}}
        \\ \cmidrule{2-7}
                        & 
                          Train & Test &
                                         Train & Test &
                                                        Train & Test
        \\ \midrule
        Laptops & 987 &  341 & 866 & 128 & 460 & 169 \\
        Restaurants & 2,164 &  728 & 805 & 196 & 633 & 196 \\
        \bottomrule
      \end{tabular}
    }
  \end{center}

  \caption{Top: Detailed statistics of the corpora for BERT language model finetuning. Bottom: Number of labels for each category of the SemEval 2014 Task 4 Subtask 2 laptop and restaurant datasets for Aspect-Target Sentiment Classification.
  }\label{tab:datasets}
\end{table}

\subsection{Hyperparameters}
\rood{Outcome: Based on the description of the hyperparamters readers should be able to reproduce our results}

We use BERT\textsubscript{BASE}\footnote{We make use of both BERT-base-uncased and XLNet-base-cased models as part of the pytorch-transformers library:  \url{https://github.com/huggingface/pytorch-transformers}} (uncased) as the base for all of our experiments, with the exception of XLNet\textsubscript{BASE} (cased), which is used as one of the baseline models.

For the BERT language model finetuning we use 32 bit floating point computations using the Adam optimizer~\cite{ArxKingma2014}. The batchsize is set to 32 while the learning rate is set to $3\cdot10^{-5}$. The maximum input sequence length is set to 256 tokens, which amounts to about 4 sentences per sequence on average.
As shown in \autoref{tab:datasets}, we finetune the language models on each domain so that the model trains a total of about 30 million sentences ($\approx$ 7.5 million sequences). 

For training the BERT and XLNet models on the down-stream task of ATSC we use mixed 16 bit and 32 bit floating point computations, the Adam optimizer, and a learning rate of $3\cdot10^{-5}$ and a batchsize of 32. We train the model for a total of 7 epochs. The validation accuracy converges after about 3 epochs of training on all datasets, but training loss still improves after that.

\begin{figure}[t!]
  \includegraphics[width=0.48 \textwidth]{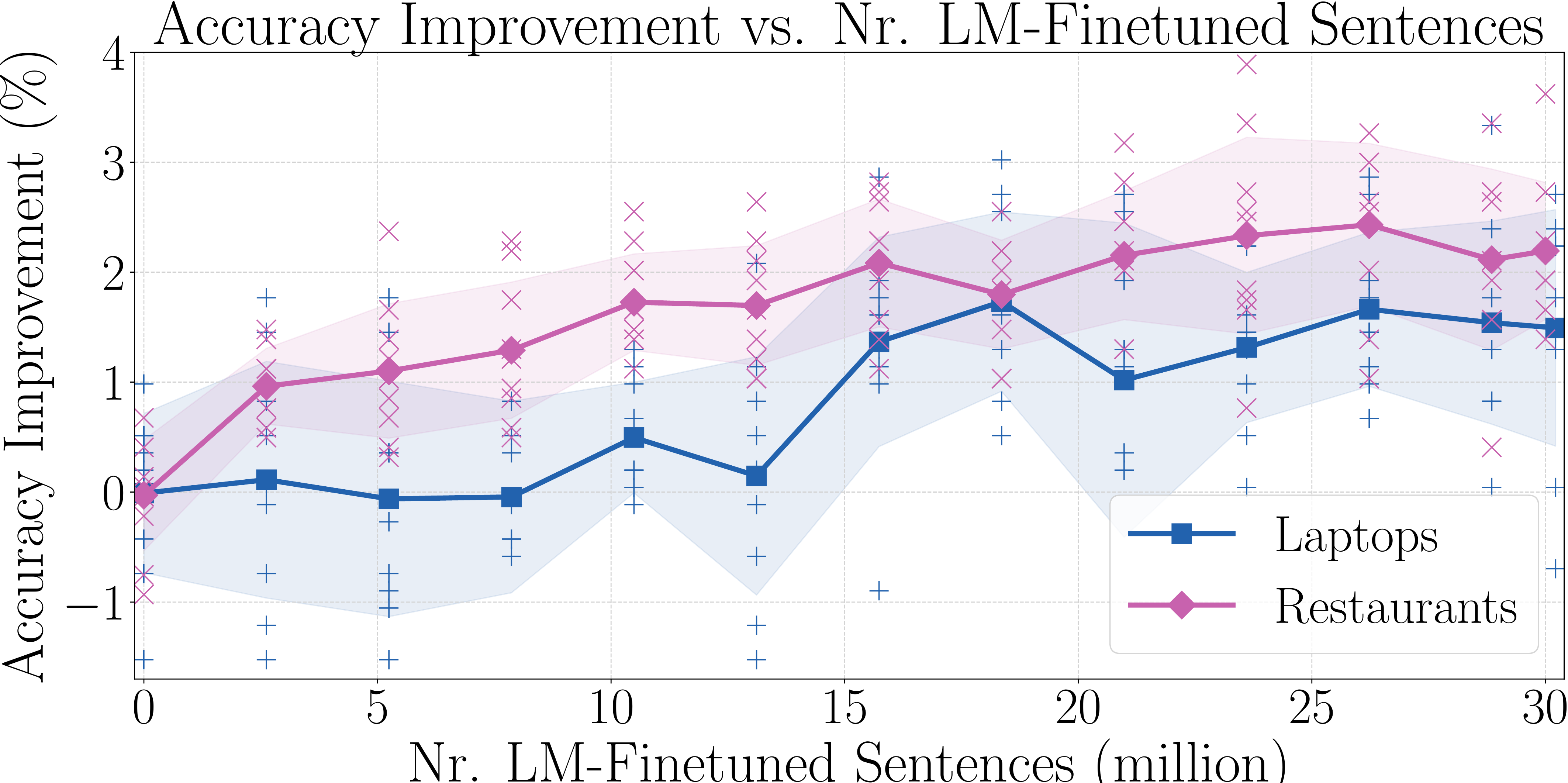}
  \caption{Absolute accuracy improvement of Aspect-Target Sentiment Classification as a function of the number of sentences the BERT language model has been finetuned on. Markers ($\blacksquare, \blacklozenge$) connected through the lines are the averages ($\mu$) over 9 runs, a single run is marked as either a cross ($\times$ for restaurants) or a plus ($+$ for laptops). The standard deviation ($\sigma$) curves are also drawn ($\mu \pm \sigma$).
    The model is trained on the SemEval 2014 Task 4 datasets and evaluated in-domain. The language models are finetuned on the target-domain corpora. Best viewed in color.} \label{fig:acc-dep-lmiterations}
\end{figure}

It is important to note that all our results reported are the average of 9 runs with different random initializations. This is needed to measure significance of improvements, as the standard deviation in accuray amounts to roughly $1\%$ for all experiments (see \autoref{fig:acc-dep-lmiterations}).

\subsection{Compared Methods}
\rood{Outcome: Readers have a short overview of the different models that are compared to our models}

We compare in-domain results to current state-of-the-art methods, which we will now describe briefly. \\
\textbf{SDGCN-BERT}~\cite{ArxZhaoa2019}
explicitly models sentiment dependencies for sentences with multiple aspects with a graph convolutional network. This method is current state-of-the-art on the SemEval 2014 laptops dataset. \\
\textbf{AEN-BERT}~\cite{ArxSong} is an attentional encoder network. When used on top of BERT embeddings this method performs especially well on the laptops dataset. \\
\textbf{BERT-SPC}~\cite{ArxSong}
is BERT used in sentence-pair classification mode. This is exactly the same method as our BERT-base baseline and therefore, we can cross-check the authors' results. \\
\textbf{BERT-PT}~\cite{Xu2019}
uses multi-task fine-tuning prior to downstream classification, where the 
BERT language model is finetuned jointly with a question answering task. It has state-of-the-art performance on the restaurants dataset prior to this paper.

To our knowledge, cross- and joint-domain training on the SemEval 2014 Task 4 datasets has not been analyzed so far. 
Thus, we compare our method to two very strong baseline models: BERT-base and XLNet-base. \\
\textbf{BERT-base}~\cite{Devlin2019}
is using the pre-trained BERT\textsubscript{BASE} embeddings directly on the down-stream task
without any domain specific language model finetuning. \\
\textbf{XLNet-base}~\cite{ArxYang2019}
is a method also based on general language model pre-training similar to BERT. Instead of randomly masking tokens for pre-training like BERT, a more general permutation objective is used, where all possible variants of masking are fully exploited.

Our models are BERT models whose language model has been finetuned on different domain corpora. \\
\textbf{BERT-ADA Lapt}
is the BERT language model finetuned on the laptop domain corpus. \\
\textbf{BERT-ADA Rest} 
is the BERT language model finetuned on the restaurant domain corpus. \\
\textbf{BERT-ADA Joint}
is the BERT language model finetuned on the corpus containing an equal amount of laptops and restaurants reviews.

\begin{table*}[h!]
\begin{center}
\scalebox{0.85}{
\begin{tabular}{lcccccc|cccccc}
\toprule
\textbf{Test Dataset} & 
\multicolumn{6}{c|}{\textbf{Laptops}} &
\multicolumn{6}{c}{\textbf{Restaurants}}
\\\midrule

\textbf{Train Dataset} & 
\multicolumn{2}{c}{\textbf{Laptops}} &  \multicolumn{2}{c}{\textbf{Restaurants}} & \multicolumn{2}{c|}{\textbf{Lapt. + Rest.}} &
\multicolumn{2}{c}{\textbf{Restaurants}} &  \multicolumn{2}{c}{\textbf{Laptops}} & \multicolumn{2}{c}{\textbf{Lapt. + Rest.}} 
\\
\textbf{Train Type} & 
\multicolumn{2}{c}{\textbf{In $\rightarrow$}} &  \multicolumn{2}{c}{\textbf{Cross $\leftrightarrow$}} & \multicolumn{2}{c|}{\textbf{Joint $\cup$}} &
\multicolumn{2}{c}{\textbf{In $\rightarrow$}} &  \multicolumn{2}{c}{\textbf{Cross $\leftrightarrow$}} & \multicolumn{2}{c}{\textbf{Joint $\cup$}} 

\\\midrule

\textbf{Other Methods} & 
\textbf{Acc} & \textbf{MF1} & 
\textbf{Acc} & \textbf{MF1} &
\textbf{Acc} & \textbf{MF1} &
\textbf{Acc} & \textbf{MF1} &
\textbf{Acc} & \textbf{MF1} & 
\textbf{Acc} & \textbf{MF1} \\

SDGCN-BERT & 
\textbf{81.35} & \textbf{78.34} & - & - & - & - &
83.57 & 76.47 & - & - & - & - \\

AEN-BERT & 
79.93 & 76.31 & - & - & - & - &
83.12 & 73.76 & - & - & - & -  \\

BERT-SPC & 78.99 & 75.03 & - & - & - & - &
84.46 & 76.98 & - & - & - & - \\

BERT-PT & 78.07 & 75.08 & - & - & - & - & 
84.95 & 76.96 & - & - & - & -
\\
\midrule

\textbf{Baselines}   \\

XLNet-base & 79.89 & 77.78 & 77.78 & 72.24  & \textbf{80.88} & 76.92 &
85.84 & 78.35 & 82.41 & 72.98 & 86.15 & 78.93 \\

BERT-base & 77.69 & 72.60 & 75.86 & 70.78 & 78.81 & 74.47 &
84.92 & 76.93 & 80.07 & 69.93 & 85.03 & 77.35 \\
\midrule

\textbf{Ours}   \\  

BERT-ADA Lapt & 79.19 & 74.18 & \lgray \textbf{77.92}  & \lgray \textbf{72.99} & 80.23 & 75.77 &
85.51 & 78.09 & 80.68  & 72.93  & 86.22 & 79.79  \\

BERT-ADA Rest & 78.60  &  74.09 & 76.16  & 70.46  & 79.14 & 74.93 &
\textbf{87.14}  & \textbf{80.05} & \lgray \textbf{83.68}  & \lgray 72.91 & \textbf{87.89} & 81.05 \\

BERT-ADA Joint & 78.96  & 74.18  & 75.91 & 69.84  & 79.94 & \textbf{78.74} &
86.35  & 78.89  & 82.23 & \textbf{73.03}  & 87.69  & \textbf{81.20} \\
\bottomrule
\end{tabular}
}
\end{center}
\caption{Summary of results for Aspect-Target Sentiment Classification for in-domain, cross-domain, and joint-domain training on SemEval 2014 Task 4 Subtask 2 datasets. 
The cells with gray background correspond to the cross-domain adaptation case, where 
the language model is finetuned on the target domain.
As evaluation metrics accuracy (Acc) and Macro-F1 (MF1) are used. }\label{tab:results}
\end{table*}

\subsection{Results Analysis}
\rood{Outcome: Readers should get a feel for why the results are how they are. 
  First describe the bigger picture in the sense like method x/y performs best.
  We think that method x/y perform best because of w/z \\
  First split results into Categories and then 
  think about all the relevant statements you want to make
}
The results of our experiments are shown in \autoref{fig:acc-dep-lmiterations} 
and \autoref{tab:results} respectively.

To answer RQ1, which is concerned with details of domain-specific language model finetuning, we can see in \autoref{fig:acc-dep-lmiterations} that first of all, language model finetuning has a significant effect on ATSC end-task performance. 
Secondly, we see that in the restaurants domain the performance starts to increase immediately, whereas in the laptops domain it takes about 10 million finetuned sentences before a significant increase can be measured.
After around 17 million sentences no significant improvement can be measured.
In addition, we find that the different runs have a high variance, which necessitates averaging over 9 runs to measure differences in model performance reliably.

To answer RQ2, which is concerned with in-domain ATSC performance, we see in \autoref{tab:results} that for the in-domain training case, our models BERT-ADA Lapt and BERT-ADA Rest achieve performance close to state-of-the-art on the laptops dataset and new state-of-the-art on the restaurants dataset with accuracies of $79.19\%$ and $87.14\%$, respectively. On the restaurants dataset, this corresponds to an absolute improvement of $2.2\%$ compared to the previous state-of-the-art method BERT-PT.
Language model finetuning produces a larger improvement on the restaurants dataset. We think that one reason for that might be that the restaurants domain is underrepresented in the pre-training corpora of BERT\textsubscript{BASE}.
Generally, we find that language model finetuning helps even if the finetuning domain does not match the evaluation domain. We think the reason for this might be that the BERT-base model is pre-trained more on knowledge-based corpora like Wikipedia than on text containing opinions. We show some evidence for this hypothesis in \autoref{sect:Casestudy}.
In addition, we find that the XLNet-base baseline performs generally stronger than BERT-base, but only outperforms the BERT-ADA models on the laptops dataset with an accuracy of $79.89\%$ .

To answer RQ3, which is concerned with domain adaptation, we can see from the grayed out cells in \autoref{tab:results}, which
correspond to the cross-domain adaption case where the BERT language model is trained on the target domain, that domain adaptation works well with $2.2\%$ absolute accuracy improvement on the laptops test set and even $3.6\%$ accuracy improvement on the restaurants test set compared to BERT-base.

In general, the ATSC task generalizes well cross-domain, with about a $2$-$3\%$ drop in accuracy compared to in-domain training. We think the reason for this might be 
that syntactical relationships between the aspect-target and the phrase expressing sentiment polarity, as well as knowing the sentiment-polarity itself, are sufficient to solve the ATSC task in most cases.

For the joint-training case, we find that combining both training datasets improves performance on both test sets.
This result is intuitive, as more training data generally leads to better performance if the domains do not confuse each other.
Interestingly, for the joint-training case the BERT-ADA Joint model
performs especially well when measured by the Macro-F1 metric. A reason for this 
might be that the SemEval 2014 datasets are imbalanced due to dominance of positive labels. 
It seems like through finetuning the language model on both domains the model learns to classify the neutral class much better, especially in the laptops domain.

\subsection{Case Study}
\label{sect:Casestudy}

\rood{Outcome: Here Readers should get a feel for how
  the finetuning on different domains and when the model is trained
  cross domain, multidomain and only on its domain.
  Maybe also show a BERT MHA visualization.
  \begin{itemize}
  \item Q1: Which world/word/domain knowledge is needed
    to correctly classify restaurant or laptop reviews?
  \item Q2: Does sentiment word-context confusions occur when mixing datasets
  \item Q3: Which kind of Sentiments are correctly classified with respect to  bert-base - in progress
  \item Q4: Which kind of generalization errors occur in out-of-domain generalization
  \end{itemize}}

The goal of the case study is to find answers to the following questions:
\begin{itemize}
 \item What specifically causes the finetuned language models BERT-ADA Lapt and BERT-ADA Rest to perform better than BERT-base?
 \item What reasons can we find by comparing conflicting predictions made by these models?
 \item What are specific reasons for erroneous classifications?
 \item What error types prevent us from performing at human expert level on ATSC?
\end{itemize}

To answer these questions we performed input reduction, which allows for a better interpretation of sample predictions from the SemEval 2014 Restaurant and Laptops test set, see \autoref{tab:casestudy}.
The input reduction technique tries to isolate a set of words from the sentence that contribute most to the prediction. The  theoretical details of input reduction are discussed in \autoref{sect:Input-reduction}.
\begin{table*}[h!]
\begin{center}
\scalebox{1}{
\begin{tabular}{lp{9.0cm}p{2.0cm}cccc}

\textbf{Ref.} & \textbf{Restaurant Samples} & \textbf{Aspect} &  \rotatebox{90}{\textbf{B}ase} & \rotatebox{90}{\textbf{L}apt} &   \rotatebox{90}{\textbf{R}est} & \rotatebox{90}{Gold} \\ \midrule

RC1 & Certainly not the \gipr{best}{R} sushi in New York, however, 
it is always fresh, and the place is very \gipr{clean}{L}, \ripr{sterile}{B}. & place & -- & + & + & + \\ 


RC2 & the \gipr{icing}{L} MADE this cake, it was \gipr{fluffy}{R}, \ripr{not}{B} ultra \ripr{sweet}{B}, creamy and light. & cake & -- & + & + & + \\ 


RC3 & The sangria's - \ipr{\gc{w}\rc{a}\gc{t}\rc{e}\gc{r}\rc{e}\gc{d}}{\gc{B,}\rc{L,R}} \ripr{down}{L}. & sangria & + & -- & -- & -- \\ 

RC4 & The staff \ripr{should}{L,R} be a bit \ripr{more}{L} \gipr{friendly}{B}. & staff & + & -- & -- & -- \\ 







RC5 & \gipr{15\%}{B} gratuity \ipr{\gc{a}\rc{u}\gc{t}\rc{o}\gc{m}\rc{a}\gc{t}\rc{i}\gc{c}\rc{a}\gc{l}\rc{l}\gc{y}}{\rc{R,}\gc{L}} added to the \ripr{bill}{R}. & gratuity & + & + & -- & -- \\ \midrule

RE1 & My friend \gripr{had}{L} a burger and I had these \gipr{wonderful}{B,R} blueberry pancakes. & burger & o & + & + & o \\

RE2 & The sauce is \gipr{excellent}{B,L,R} (very fresh) with dabs of real mozzarella. & dabs of real mozzarella & + & + & + & o \\ \midrule

 & \textbf{Laptop Samples} & &  & & & \\ \midrule


LC1 & the retina display display make pictures \gipr{i}{R} \ipr{\gc{t}\rc{o}\gc{o}\rc{k}}{\rc{B,}\gc{L}} \ripr{years}{B} ago \gipr{jaw}{R} dropping. & retina display display & -- & + & + & + \\

LC2 & The Mac mini is about 8x smaller than my old computer which is a huge bonus and \ripr{runs}{B} \ipr{\gc{v}\rc{e}\gc{r}\rc{y} \gc{q}\rc{u}\gc{i}\rc{e}\gc{t}}{\rc{B,}\gc{L}}, \ripr{actually}{B} the fans \ripr{aren't audible}{R} unlike my old pc & fans & -- & + & -- & + \\ \midrule


LE1 & the latest version does \ripr{not}{B,R,L} have a disc drive. & disc drive & -- & -- & -- & o \\ 

LE2 &  Which it did \ripr{not}{B,R} have, \ripr{only}{L} 3 USB 2 ports. & USB 2 ports & -- & -- & -- & o \\


\end{tabular}
}
\end{center}
\caption{
Shown are text samples from SemEval 2014 Restaurants and Laptops test-set that are  predicted correctly for the language model adapted to the target domain but  predicted falsely with the bert-base model (RC1-5, LC1-LC2). In addition, samples which are  predicted falsely by the target-domain adapted model are shown (RE1-2, LE1-2).
The abbrevations stand for: B -- BERT-base, L -- BERT-ADA Lapt(op), R -- BERT-ADA Rest(aurant) -- all the language models used for prediction. The used down-stream-classifiers are trained in-domain.
The reduced input (set of words that influence prediction strongest) is formatted with underline and the subscript denotes the corresponding model (B, L, R) used for computing the reduced input. If viewed in color, the corresponding predicted sentiment polarity of the reduced input corresponds to: green -- positive, red -- negative, gray -- neutral, alternating green and red -- both negative and positive for different models.
Best viewed in color.
}\label{tab:casestudy}
\end{table*}
\subsubsection*{Samples predicted correctly by the target-domain adapted model}

In the following, we will discuss a selection of examples that are classified correctly by the best performing in-domain BERT-ADA  and incorrectly by BERT-base.
The error types for BERT-base are mentioned for all the examples next to their reference label.
\\
\textbf{RC2} -- restaurant-domain context needed:  \\
\textit{``not \dots sweet''} -- the negated sentiment contained in this phrase is identified correctly by BERT-base and could be interpreted as negative in isolation, but an adjective like \textit{``fluffy''} carries a stronger positive sentiment for BERT-ADA Rest. 
\\
\textbf{RC4} -- general review-domain context needed: \\ 
We think that \textit{``should be''} is an expression often found in text with opinions, thus both BERT-ADA Lapt and Rest, which both have been finetuned on review-specific text, predict this example correctly. BERT-base is strongly influenced by \textit{``friendly''} and cannot detect the sentiment-negating function of \textit{``should be''}. 
\\
\textbf{RC5} -- restaurant-domain context needed: \\ 
The reduced input \textit{``gratuity''} is detected as positive for the BERT-ADA Laptop and BERT-base model. In contrast, the BERT-ADA Rest model reveals reduced input words \textit{``automatically''} and \textit{``bill''} to detect the negative sentiment correctly. 
\\
\textbf{LC2} -- laptop-domain context needed: \\
\textit{``very quiet''} is classified as negative by BERT-base whereas the same expression is classified positive by the BERT-ADA Lapt model. BERT-ADA Rest classifies ``aren't audible'' as negative.

\subsubsection*{Samples predicted incorrectly by the target-domain adapted model}

In the following, we investigate examples that are classified incorrectly by the BERT-ADA models. This helps us to understand 
the remaining error types and shows a way forward for future work.
The majority of incorrect predictions come from the ground-truth neutral class, which in most cases is confused with the positive class for restaurants and with the negative class for laptop reviews. 
\\
\textbf{RE1} -- influenced by sentiment towards a different aspect-target: \\
This examples was classified correctly only by the BERT-ADA Laptop model. The reduced input for this model is the word \textit{``'had''}, which is used a lot in fact based formulations like for example \textit{``the CPU had 3 GHz''}. From experience, we think that this type of formulation appears more often in the laptops than in the restaurant domain. The BERT-ADA Restaurant and BERT-base model both seem to be influenced by the sentiment associated to another aspect-target.
\\
\textbf{RE2} -- influenced by sentiment towards a different aspect-target: \\
Words indicating a certain kind of relation to the aspect-target like ``with''  in this example could be used to separate the aspect-target specific sentiment from the general sentiment. We think that with more supervised data this case should be solvable by learning these relations in a general way.
\\
\textbf{LE1} -- absence of something like a part classified as negative: \\
\textit{``not''} is classified as negative by BERT-ADA Lapt. In the laptops domain the largest remaining confusion are neutral examples classified as negative examples by our algorithm. It seems if absences of parts like a ``disk drive'' are mentioned, the algorithm tends to classify this as negative. In other examples these statements of absence of things actually imply a negative sentiment.
\\
\textbf{LE2} -- possibly incorrect ground truth: \\
A handful of examples like this one are, in our opinion, labelled incorrectly. We think the word \textit{``only''} indicates negative sentiment in this examples.
  
\rood{Types: context specific sentiment inverters (gratuity added to bill vs. gave gratuity to waiter); general sentiment negators (should); dissocation from general sentiment; context-specific interpretation of a quantitative formulation (has no dvd drive, 5 min away location, 5-dishes menu, 4 usb ports, 5 Ghz processor) of an aspect-target}

To summarize, we find that in order to correctly predict aspect-target based sentiment, 
the context sensitivity of the sentiment expression plays a important role in difficult examples. 
By finetuning the language model on domain-specific text the model is able to capture this knowledge most of the time, even if such expressions are not directly observed in the training set used for downstream-classification.

We see that especially neutral examples are more difficult to classify correctly. Some of these examples could be solved for an applied real-world case with more supervised data that allows to learn more abstract relationships between entities like sauce and its ingredients in example RS7 and contain more fact-based formulations to discriminate the neutral class better.
We also think that selecting finetuning corpora more carefully with these error types in mind could also lead to improvements of classification performance on these datasets.

\section{Conclusion}
\label{sect:Conclusion}
\rood{Abstract in Past-Tense without motivation}

We performed experiments on the task of Aspect-Target Sentiment Classification
by first finetuning a pre-trained BERT model on a domain specific corpus with subsequent training on the down-stream classification task.

We analyzed the behavior of the number of domain-specific BERT language model finetuning steps in relation to the end-task performance.

With the findings on how to best exploit BERT language model finetuning we were able
to train high performing models, of which the one trained on SemEval 2014 Task 4 restaurants dataset achieves new state-of-the-art performance.

We further evaluated our models cross-domain to explore the robustness of Aspect-Target Sentiment Classification. We found that with our setup, this task transfers 
well between the laptops and the restaurants domain.

As a special case we ran a cross-domain adaptation experiments, where
the BERT language model is specifically finetuned on the target domain.
We achieve significant improvement over unadapted models: one cross-domain adapted model performs even better than a BERT-base model that is trained in-domain.

Overall, our findings reveal promising directions for follow-up work. The XLNet-base model performs strongly on the ATSC task. Here, domain-specific finetuning could probably bring significant performance improvements. 
Another interesting direction for future work would be to investigate cross-domain behavior for an additional domain like hotels, which is more similar to the restaurants domain.

\bibliography{references}
\bibliographystyle{acl_natbib}

\appendix

\end{document}